\documentclass{IOS-Book-Article}

\usepackage{mathptmx}
\usepackage{soul}\setuldepth{article}
\usepackage{amsmath,amssymb}
\usepackage{booktabs}   
\usepackage[nocompress]{cite}
\usepackage{url}
\usepackage{booktabs}
\usepackage[table]{xcolor}

\def\hb{\hbox to 11.5 cm{}}

\newcommand{\TauStrict}{0.28}
\newcommand{\TauBal}{0.20}

\begin{document}

\pagestyle{headings}
\def\thepage{}

\begin{frontmatter}

\title{Retrieving Semantically Similar Decisions under Noisy Institutional Labels: Robust Comparison of Embedding Methods}

\markboth{}{September 2025\hb}

\author[A]{\fnms{Tereza} \snm{Novotna}\orcid{0000-0002-1426-4547}\thanks{Corresponding author: Tereza Novotna, email: tereza.novotna@law.muni.cz}},
and
\author[B]{\fnms{Jakub} \snm{Harasta}\orcid{0000-0002-5722-0325}}

\runningauthor{Novotna}

\address[A]{Faculty of Law, Masaryk University, Brno, Czechia}
\address[B]{Faculty of Law, Masaryk University, Brno, Czechia}

\begin{abstract}
Retrieving case law is a time-consuming task predominantly carried out by querying databases. We provide a comparison of two models in three different settings for Czech Constitutional Court decisions: (i) a large general-purpose embedder (OpenAI), (ii) a domain-specific BERT-trained from scratch on $\sim$30{,}000 decisions using sliding windows and attention pooling. We propose a noise-aware evaluation including IDF-weighted keyword overlap as graded relevance, binarization via two thresholds (\TauBal\ balanced, \TauStrict\ strict), significance via paired bootstrap, and an nDCG diagnosis supported with qualitative analysis. Despite modest absolute nDCG (expected under noisy labels), the general OpenAI embedder decisively outperforms the domain pre-trained BERT in both settings at @10/@20/@100 across both thresholds; differences are statistically significant. Diagnostics attribute low absolutes to label drift and strong ideals rather than lack of utility. Additionally, our framework is robust enough to be used for evaluation under a noisy gold dataset, which is typical when handling data with heterogeneous labels stemming from legacy judicial databases.

\end{abstract}


\end{frontmatter}

\markboth{September 2025\hb}{September 2025\hb}

\section{Introduction}

Searching for similar decisions constitutes a large part of the work of analysts, clerks, and judges at the Constitutional Court (ConC). We evaluate the possibility to deploy automated methods to support the retrieval of semantically similar decisions.

Even though Czech jurisdiction is not precedent-based, legal certainty dictates awareness of previous case law in similar legal matters. In the Czech (continental) legal system, citing prior case law is not formally regulated, i.e., there are few to no rules regarding the similarity of facts or legal issues. Therefore, judges may cite decisions for various reasons and with varying degrees of similarity. The informality influences the understanding of \textit{similarity}, and the choice of experimental and evaluation methods to engage in retrieving semantically similar decisions from ConC database.



Czech ConC provides public access to its case law through the database NALUS\footnote{nalus.usoud.cz}. Standard retrieval techniques include querying the database for cases related to specific legal provisions and searching cases with similar keywords. The list of keywords consists of 713 items assigned to individual ConC decisions based on their thematic relevance. Keywords are assigned manually by the ConC's staff, which is why we consider it to be a gold dataset. However, given the rotation of ConC's staff, the lack of formal rules, and the rise of new legal issues, the dataset is noisy, marked by inconsistent methodology in assigning keywords across time and topics. The fact is well-known in Czechia, but its impact (e.g., implications for automation) understudied.

We compare two models (domain-trained BERT and off-shelf general OpenAI embedder) in three embedding-based retrieval methods on a large corpus and introduce an evaluation that tolerates noisy gold: IDF-weighted keyword overlap as graded gains; two relevance thresholds (\TauBal, \TauStrict) calibrated from data; paired bootstrap for significance; and nDCG diagnostics.

The Introduction is followed by Section 2 containing related work. Section 3 provides description of Data and Benchmark, and Section 4 outlines the methods. We report results in Section 5 together with qualitative example of the noisiness. In Section 6, we discuss our findings, and provide Conclusion and Future Work in Section 7.

We provide the following contributions to AI\&Law: (1) robust reproducible framework for LIR under noisy taxonomies, (2) study showing a large general embedder $\gg$ domain BERT trained from scratch on $\sim$30k decisions, and (3) diagnostics (label drift, per-bin slices, decomposition) explaining modest absolute nDCG while preserving strong relative differences

\section{Related Work}

Given that the legal domain is predominantly expressed through written language, it is well-suited for the application of NLP and ML methods and tools \cite{bench2012history, governatori2022thirty, sartor2022thirty, villata2022thirty, katz2023natural, siino2025exploring}. Significant advances in law-related NLP were brought by pre-trained Transformer-based Language Models (TLMs) and subsequent introduction of Large Language Models (LLMs) to the public \cite{greco2024bringing, lai2024large, siino2025exploring, naveed2025comprehensive}. We are continuing this trend of applying NLP in the legal domain.

Legal information retrieval represents a long-standing research area with both theoretical challenges and practical implications. Text embeddings and TLMs/LLMs \cite{mikolov2013efficient, le2014distributed, devlin2019bert} are regularly implemented to enhance document retrieval. Models pre-trained for use in specific domains, such as SciBERT \cite{beltagy2019scibert} for scientific literature, BioBERT \cite{lee2020biobert} for biomedical literature, or Legal-BERT \cite{chalkidis2020legal} for law, contribute to increased performance for domain-specific tasks. In case law retrieval, efforts include unsupervised techniques \cite{mandal2021unsupervised} or inclusion of supporting tools and models into the retrieval pipeline \cite{westermann2020paragraph,vuong2023sm,mentzingen2025effectiveness,tang2025uqlegalai}. We engage in the retrieval of the Czech ConC case law by using and evaluating the use of text embeddings.

LLMs are deployed for a plethora of domain-specific tasks where their zero-shot and few-shot capabilities offer added value \cite{savelka2023unreasonable, lee2025long, dugac2025classifying}. Limitations related to various models and their use are frequently reported \cite{dahl2024large, magesh2025hallucination}, with increasing effort going into prompt-engineering techniques and RAG to mitigate shortcomings in the high-stakes domain \cite{yu2023exploring, zin2024leveraging, keisha2025all}. Contribution to various known bottlenecks, e.g., annotation \cite{savelka2023unreasonable}, thematic analysis \cite{drapal2023using}, legal aid intake \cite{steenhuis2024getting} or law-related self-help \cite{gogani2025performance}, is well-established. Evaluation of performance is still challenging, given the scarcity of complex domain-specific benchmarks akin to \cite{guha2023legalbench}. Comparative evaluation efforts focusing on plethora of tasks and methods similar to \cite{guo2025specialized} appeared only recently. We follow suit, employing OpenAI embedder and providing robust comparative evaluation of its performance on real-world data and well-known domain-specific tasks.

\section{Data and Benchmark}
\subsection{Corpus and Metadata}
We used a subset of ConC decisions from the Czech Court Decisions Corpus (CzCDC) \footnote{hdl.handle.net/11372/LRT-3052}. The subset contains 73,086 ConC decisions from January 1993 to September 2018. Subsequently, we retrieved data from ConC's NALUS, and obtained 18,238 decisions from October 2018 to September 2024. In total, there are 91,324 decisions. We also obtained all the metadata contained in this database: file number, decision and publication dates, reporting judge, related legal provisions, and keywords (in orig. \textit{"Věcný rejstřík"}). These are all the decisions of this court since its establishment in January 1993.

After discussion with ConC's analytical department and inspection of the dataset of all decisions, we selected only decisions longer than 10,000 characters. Decisions shorter than this benchmark do not contain substantive legal arguments and are only formal rejections of constitutional complaints on procedural grounds (e.g., failure to meet the deadline). After this removal, 34,441 decisions remained, which we used for training BERT model and then for creating vector representations for all methods.

\subsection{Label Drift and Incompleteness}


First, analysed and described the keywords dataset to determine how it might influence the evaluation. We calculated the annual average number of keywords per decision, the number of unique keywords, the median number, the ratio of decisions with zero keywords, and the entropy of keywords. Due to the scope of the article, we present only the relevant sections below and in Table \ref{tab:overlap-sanity}.

\begin{table}[ht]
\caption{Back-of-the-envelope overlap under diversified labels.}
\label{tab:overlap-sanity}
\centering
\begin{tabular}{lrrrrr}
\toprule
Scenario & $k$ & $H$ (bits) & $N_{\text{eff}}$ & $\mathbb{E}[|A\cap B|]$ & $P(\text{zero})$ \\
\midrule
Typical year        & 3 & 6.9 & 118 & 0.076 & 0.93 \\
Richer year         & 3 & 7.8 & 226 & 0.040 & 0.96 \\
Aggressive tagging  & 5 & 6.9 & 118 & 0.212 & 0.81 \\
\bottomrule
\end{tabular}
\end{table}

\emph{Typical year} uses the corpus medians: $k$ is the median number of keywords per decision and $H$ is the median annual doc-frequency entropy (bits), with $N_{\text{eff}}=2^{H}$ the effective active vocabulary.
\emph{Richer year} takes a high-diversity regime (e.g., upper-quartile $H$ by year); $k$ is the median for that year (or kept equal to Typical to isolate the effect of diversity). This shows that as the label space becomes larger and more uniformly used, the expected exact overlap $k^2/N_{\text{eff}}$ decreases.
\emph{Aggressive tagging} holds the vocabulary regime fixed (same $H$ as Typical) but increases $k$ (e.g., the 75th percentile or a policy of $\approx\!5$ tags per decision). This improves overlap but, when $N_{\text{eff}}$ is large, $k^2/N_{\text{eff}}\!\ll\!1$ so $P(|A\cap B|=0)\approx e^{-k^2/N_{\text{eff}}}$ remains high.

Our diagnostics show that, over time, decisions receive slightly more keywords, but the label space simultaneously becomes larger and more uniformly used (higher entropy). Exact keyword overlap between two similar decisions scales roughly with $k^2/N_{\text{eff}}$, where $k$ is the number of keywords per decision and $N_{\text{eff}}=2^{H}$ is the effective size of the active vocabulary implied by the doc-frequency entropy $H$. With typical values $k\approx 3$ and $N_{\text{eff}}\approx 100\text{--}200$, the expected overlap $\mathbb{E}[|A\cap B|]$ is $\ll 1$ and the probability of zero overlap $P(|A\cap B|=0)\approx \exp(-k^2/N_{\text{eff}})$ remains high (often $\ge 0.8$). Consequently, overlap-based gains - and therefore absolute nDCG - are conservative lower bounds on semantic quality in this corpus, even though relative differences between models are large and statistically significant.


\subsection{Query Set and Candidate Pool}

We form a query set of 2,000 decisions with assigned keywords (from metadata) from the dataset. To ensure that keyword overlap is at least measurable, we restrict the sampling pool to decisions with \(\ge 2\) keywords. We then apply stratified sampling along three axes that affect evaluation difficulty: 
\begin{enumerate}
\item Keyword count per decision, binned as \(\,2\text{--}3,\ 4\text{--}7,\ 8+\).
\item Rarity of a decision’s tags, computed as the \emph{median} IDF of its (deduplicated) keyword set (IDF defined below) and split at the median into \emph{common} vs.\ \emph{rarer}.
\item Decision year, obtained from "Date of publication" metadata, and binned by empirical quantiles (three bins; ties handled by \texttt{qcut} with duplicates allowed).
\end{enumerate}
Within the resulting \(3\times 2\times 3=18\) strata, we allocate samples approximately evenly with a fixed random seed (reallocating surplus to larger strata when a stratum is under-populated). Each sampled query is identified by its normalized document ID.

Subsequently, at retrieval time, each model searches its full indexed corpus (we do not restrict the candidate set beyond the model’s own coverage), returns the top-\(k\) neighbors by cosine similarity with unit-normalized vectors, and excludes the self-match. Unless noted otherwise, \(k=100\) and we report metrics at \(k\in\{10,20,100\}\).

\subsection{Graded Relevance via IDF-Weighted Overlap}

Let \(A\) and \(B\) be the keyword sets for two decisions. We compute inverse document frequency over \emph{keyworded} decisions as
\[
\mathrm{IDF}(t) \,=\, \log\!\frac{N}{df(t)},
\]
where \(df(t)\) is the number of decisions whose keyword set contains \(t\), and \(N\) is the number of decisions with at least one keyword.\footnote{We preserve slashes inside a term (e.g., \emph{``právní domněnka/presumpce''}) and do not split such multiword forms; this respects the court’s taxonomy and avoids artificial fragmentation.} 

We take as \emph{graded gain} the IDF-weighted Jaccard similarity:
\begin{equation}
\label{eq:wjacc}
\mathrm{wJacc}(A,B)
= \frac{\sum_{t\in A\cap B}\mathrm{IDF}(t)}{\sum_{t\in A}\mathrm{IDF}(t)+\sum_{t\in B}\mathrm{IDF}(t)-\sum_{t\in A\cap B}\mathrm{IDF}(t)} \;\in [0,1].
\end{equation}
This rewards overlap on \emph{diagnostic} (high-IDF) keywords and down-weights ubiquitous keywords. We use \(\mathrm{wJacc}\) directly as the graded relevance \(r_i\) at rank \(i\) for nDCG and also report binarized metrics (P@\(k\), MAP@\(k\), RBP) by thresholding \(r_i\) at two data-driven levels \(\tau\in\{\TauBal,\ \TauStrict\}\). The \emph{Balanced} track (\(\tau=\TauBal\)) captures moderate overlaps; the \emph{Strict} track (\(\tau=\TauStrict\)) focuses on high-confidence overlaps. Thresholds are chosen by inspecting the empirical percentile distributions of \(\mathrm{wJacc}\) for retrieved neighbors and for the corpus-level “ideal” candidates (i.e., all decisions sharing at least one keyword), and placed near the knees of those curves.\footnote{Operationally, nDCG uses the true ideal (IDCG) computed from \(\mathrm{wJacc}\); queries with IDCG@\(k\!=\!0\) are excluded from nDCG aggregates at that \(k\) but are retained for binarized metrics via their \(R_\tau\) denominators.}
We don't report Recall@$k$ because the denominator $R_\tau$ (all $\tau$-relevant items in the full corpus) is large, making Recall@$k$ numerically small even for systems with strong head quality. In our setting, graded nDCG and head-oriented P@$k$/HitRate@$k$/RBP better reflect user utility.

\section{Methods}

\subsection{BERT Pretraining (From Scratch, Czech Constitutional Court Corpus)}
\label{sec:bert-pretrain}

We pretrain a BERT-style encoder from scratch on the full text of \(\sim\)30k Czech Constitutional Court decisions longer than 10,000 characters in text.
Markup and boilerplate are stripped while preserving punctuation, numerals, and Czech diacritics. Text is normalized to NFC and kept \emph{cased} (we do not lowercase Czech).

We learn a subword vocabulary on the court corpus using SentencePiece (Unigram) with a 32k inventory,\footnote{Any comparable subword learner is acceptable; we preserve diacritics and do not apply additional heuristic splitting beyond whitespace and punctuation rules internal to SentencePiece.}
which is well-suited to Czech morphology.
All subsequent inputs use this tokenizer.

We use a BERT-base configuration (12 transformer layers, hidden size 768, 12 attention heads, GELU activations, layer/dropout 0.1).
Parameters are randomly initialized and trained from scratch (no initialization from generic multilingual checkpoints).

To mitigate the 512-token limit, each document is chunked into fixed-length sequences of 512 subword tokens using a sliding window with overlap (stride \(s\), e.g., \(s{=}256\)). This policy is used both during pretraining and later when producing document embeddings, ensuring that long decisions are fully covered and that local context is preserved across windows.

We train with masked language modelling (MLM) only.
Following standard practice, 15\% of tokens are selected for prediction with the 80/10/10 replacement rule (mask/keep/random) and dynamic masking per epoch.
We do not use next-sentence prediction (NSP). We trained the model on 10 epochs over 7 days on a single GPU.

At inference time, each 512-token window is encoded to hidden states \(\{h_1,\dots,h_T\}\) (we exclude \([\mathrm{CLS}], [\mathrm{SEP}]\) from pooling).
We experiment with two pooling schemes:

\begin{enumerate}
\item \textbf{Mean pooling} (simple, stable): \(e_{\text{win}}=\frac{1}{T}\sum_{i=1}^T h_i\).
\item \textbf{Self-attention pooling} (diagnostic token weighting): \(\alpha_i \propto \exp(w^\top h_i)\), \(e_{\text{win}}=\sum_i \alpha_i h_i\), with a learnable vector \(w\) (no supervision beyond the encoder; \(w\) is optimized jointly or as a light head).
\end{enumerate}

For an entire decision with windows \(W\), we aggregate window embeddings by averaging:
\(e_{\text{doc}}=\frac{1}{|W|}\sum_{w\in W} e_{\text{win}}^{(w)}\), then L2-normalize to unit length for cosine search (more in Section \ref{retrieval-setup}).

In the main comparative tables in Section \ref{results} we report the \emph{attention-based-pooled} variant for both methods as it was more accurate than average pooling (the data to prove this claim are beyond the scope of the paper but consistent with the current literature, e.g. \cite{er2016attention}).
No supervised IR fine-tuning is used. Retrieval relies purely on cosine similarity in the learned embedding space.

\subsection{OpenAI (general-purpose) model}

We use OpenAI’s \texttt{text-embedding-3-large} as an off-the-shelf embedder. The model produces 3{,}072-dimensional vectors by default and is the current high-accuracy embedding family from OpenAI, with documented gains over prior \texttt{ada-002} on MIRACL and MTEB benchmarks. It also supports an optional \emph{dimensions} parameter to shorten embeddings (for storage/speed) while retaining most retrieval quality. In our pipeline, we embed the decision’s factual section, and we report results under the label \emph{OpenAI-embedder\_facts}.

\subsection{Document scope settings}
We build two BERT embedding variants that differ only in the \emph{text span} fed to the encoder.
(\emph{i}) Full decisions: the complete opinion is chunked with a 512-token sliding window (stride 128), each window is encoded, and window embeddings are averaged into a single document vector (then L2-normalized).
(\emph{ii}) Facts only: we extract the contiguous segment that describes the factual background of the case (parties and salient events) and embed \emph{only} this segment with the same windowing/aggregation policy. The facts section is identified with simple, deterministic rules over the court’s internal layout: regexes anchored to section headings and typographic markers in the source (e.g., bold/uppercase labels, colon-terminated headers), followed by span capture until the next major heading.
The choice is motivated by the hypothesis that compressing long, multi-topic decisions into one vector induces \emph{topic averaging} and noise. In contrast, the factual description is shorter, denser, and more comparable across cases—hence more discriminative for semantic similarity.

For the OpenAI embedder, we likewise adopt the \emph{facts-only} scope in this experiment: it aligns with our hypothesis about comparability and, pragmatically, reduces token costs while allowing an entire facts segment to fit as a single input.%
\footnote{Using the full decision with the OpenAI embedder yields substantially longer inputs and higher cost without changing the retrieval pipeline. We therefore reserve that setting for future work.}

\subsection{Retrieval Setup}
\label{retrieval-setup}
Decision embeddings in all three methods are L2-normalized, so cosine similarity reduces to inner product: for a query decision with vector $e_q$ and a candidate $e_d$, $\cos(e_q,e_d)=e_q^\top e_d$.
We issue the query using the decision’s own embedding and retrieve the top-$k$ nearest neighbours ($k\in\{10,20,100\}$), excluding the self-match. 
Search is exact inner-product kNN using FAISS \texttt{IndexFlatIP} in float32.
No lexical constraints, metadata filters, or re-ranking are applied; candidates can come from any year/panel. 
For fairness across models, the \emph{query set} is the intersection of documents present in all indices, but each model searches its \emph{full} indexed corpus (we do not restrict the candidate pool to the intersection unless stated).
Ties are broken deterministically by (similarity, document ID).
All reported runs are reproducible via fixed random seeds for query sampling and stable index construction.
Intuitively, a higher cosine indicates greater directional alignment (more similar decisions).

\subsection{Evaluation Metrics and Significance}

Let the ranked list for a query be $(d_1,\ldots,d_k)$. We use a \emph{graded} gain
$r_i=\mathrm{wJacc}(A_{\text{query}},A_{d_i})\in[0,1]$ (Eq.~\ref{eq:wjacc}).
For binary metrics we threshold $r_i$ at $\tau$ and write
$b_i=\mathbf{1}[\,r_i\ge\tau\,]$.

\paragraph{nDCG@k (graded, $\tau$-independent).}
Discounted cumulative gain at $k$ and its normalization by the ideal value $\mathrm{IDCG}@k$ (sorting by $r_i$):
\begingroup\setlength{\abovedisplayskip}{4pt}\setlength{\belowdisplayskip}{4pt}
\[
\mathrm{DCG}@k=\sum_{i=1}^{k}\frac{2^{r_i}-1}{\log_2(i+1)},\qquad
\mathrm{nDCG}@k=\frac{\mathrm{DCG}@k}{\mathrm{IDCG}@k}.
\]
\endgroup
It jointly reflects how many strong overlaps exist and how early
they are ranked. Because $r_i$ is graded, nDCG does \emph{not} depend on $\tau$.
In our corpus, absolute values are conservative due to sparse/fragmented labels,
but \emph{relative} differences across models are robust.

\paragraph{P@k (binary at $\tau$).}
Precision at $k$ is the fraction of the top-$k$ that meets the threshold:
\begingroup\setlength{\abovedisplayskip}{4pt}\setlength{\belowdisplayskip}{4pt}
\[
\mathrm{P}@k=\frac{1}{k}\sum_{i=1}^{k} b_i.
\]
\endgroup
It answers a user question: ``If I look at the first $k$, how many are strong
matches?'' Higher $\tau$ \(\Rightarrow\) stricter matches \(\Rightarrow\) lower P@k.

\paragraph{MAP@k (binary at $\tau$).}
Average precision at $k$ averages the precision observed at each hit:
\begingroup\setlength{\abovedisplayskip}{4pt}\setlength{\belowdisplayskip}{4pt}
\[
\mathrm{AP}@k=\frac{1}{\sum_{i=1}^{k} b_i}\sum_{i=1}^{k} b_i\cdot\frac{\sum_{j=1}^{i} b_j}{i},
\]
\endgroup
with the convention $\mathrm{AP}@k=0$ if there is no hit in the top $k$.
We report the mean over queries (\emph{MAP}). MAP rewards systems that place
\textbf{many} relevant items \textbf{early}. Values can be small when the
relevance pool is large or when $\tau$ is stringent.

\paragraph{HitRate@k (binary at $\tau$).}
Hit rate (a.k.a.\ success@k) is the probability of finding \emph{at least one}
relevant item in the top-$k$:
\begingroup\setlength{\abovedisplayskip}{4pt}\setlength{\belowdisplayskip}{4pt}
\[
\mathrm{HitRate}@k=\mathbf{1}\!\left[\sum_{i=1}^{k} b_i \ge 1\right]
\]
\endgroup
(averaged over queries). It captures a ``first useful result'' experience and is easy to interpret.

\paragraph{RBP@10 (binary at $\tau$, user patience $p$).}
Rank-Biased Precision models a user who continues from rank $i$ to $i{+}1$ with
probability $p$:
\begingroup\setlength{\abovedisplayskip}{4pt}\setlength{\belowdisplayskip}{4pt}
\[
\mathrm{RBP}@10=(1-p)\sum_{i=1}^{10} p^{\,i-1} b_i.
\]
\endgroup
With $p{=}0.9$ we emphasize early ranks but still credit deeper results.
RBP is an expected-utility view of head quality.

\paragraph{WeightedOverlap@k and OverlapCount@k (diagnostics; $\tau$-independent).}
These are \emph{additive} summaries of the overlap in the top-$k$:
\begingroup\setlength{\abovedisplayskip}{4pt}\setlength{\belowdisplayskip}{4pt}
\[
\mathrm{OverlapCount}@k=\sum_{i=1}^{k}\!\bigl|A_{\text{query}}\cap A_{d_i}\bigr|,\qquad
\mathrm{WeightedOverlap}@k=\sum_{i=1}^{k}\ \sum_{t\in A_{\text{query}}\cap A_{d_i}}\!\mathrm{IDF}(t).
\]
\endgroup
They indicate how many keywords—and how \emph{diagnostic} (high-IDF)—are shared across
the retrieved head. They help explain nDCG movements and are useful when labels are
diverse: larger weighted overlap signals semantically closer neighbors even when
binary metrics at a high $\tau$ are conservative.

To assess statistical significance, we use paired bootstrap over queries (resample queries with replacement; report mean $\Delta$, 95\% CI, and two-sided bootstrap $p$).

\section{Results}
\label{results}
\subsection{Main Comparison: Balanced and Strict Tracks}


\begin{table}[ht]
\caption{Results at $k=10$ with $\tau=0.20$ over 2{,}000 queries. 
nDCG uses graded gains ($r=\mathrm{wJacc}$) and is $\tau$-independent.}
\label{tab:k10-tau020}
\centering
\setlength{\tabcolsep}{4pt}\footnotesize
\begin{tabular}{lcccccc}
\toprule
Model & nDCG@10 & MAP@10 & P@10 & HitRate@10 & RBP@10 & Weighted/Cnt@10 \\
\midrule
OpenAI-embedder\_facts            & \textbf{0.190} & \textbf{0.00291} & \textbf{0.260} & \textbf{0.716} & \textbf{0.175} & \textbf{42.80 / 12.18} \\
BERT\_ab\_all           & 0.090          & 0.00078          & 0.106          & 0.467          & 0.071          & 27.81 / 8.57 \\
BERT\_ab\_facts & 0.060          & 0.00027          & 0.067          & 0.364          & 0.046          & 14.28 / 4.81 \\
\bottomrule
\end{tabular}
\end{table}

\begin{table}[ht]
\caption{Results at $k=10$ with $\tau=0.28$ (more stringent relevance). 
nDCG is unchanged; binary metrics reflect the higher threshold.}
\label{tab:k10-tau028}
\centering
\setlength{\tabcolsep}{4pt}\footnotesize
\begin{tabular}{lcccccc}
\toprule
Model & nDCG@10 & MAP@10 & P@10 & HitRate@10 & RBP@10 & Weighted/Cnt@10 \\
\midrule
OpenAI-embedder\_facts            & \textbf{0.190} & \textbf{0.00501} & \textbf{0.168} & \textbf{0.550} & \textbf{0.113} & \textbf{42.80 / 12.18} \\
BERT\_ab\_all           & 0.090          & 0.00115          & 0.060          & 0.315          & 0.040          & 27.81 / 8.57 \\
BERT\_ab\_facts & 0.060          & 0.00042          & 0.038          & 0.231          & 0.026          & 14.28 / 4.81 \\
\bottomrule
\end{tabular}
\end{table}

\begin{table}[ht]
\caption{Results at $k=20$ (graded nDCG; binary metrics at $\tau\in\{0.20,0.28\}$).}
\label{tab:k20-tau020-028}
\centering
\setlength{\tabcolsep}{3pt}\footnotesize
\begin{tabular}{lcccc}
\toprule
Model & nDCG@20 & P@20 (.20/.28) & Hit@20 (.20/.28) \\
\midrule
OpenAI-embedder\_facts        & \textbf{0.187} & \textbf{0.236}/\textbf{0.151} & \textbf{0.802}/\textbf{0.649} \\
BERT\_ab\_all & 0.090        & 0.096/0.053                   & 0.600/0.406                   \\
BERT\_ab\_facts         & 0.059          & 0.058/0.032                   & 0.472/0.309                   \\
\bottomrule
\end{tabular}
\end{table}

\begin{table}[ht]
\caption{Results at $k=100$ (graded nDCG; binary metrics at $\tau\in\{0.20,0.28\}$).}
\label{tab:k100-tau020-028}
\centering
\setlength{\tabcolsep}{3pt}\footnotesize
\begin{tabular}{lcccc}
\toprule
Model & nDCG@100 & P@100 (.20/.28) & Hit@100 (.20/.28) \\
\midrule
OpenAI-embedder\_facts      & \textbf{0.182} & \textbf{0.179}/\textbf{0.109} & \textbf{0.936}/\textbf{0.827} \\
BERT\_ab\_all & 0.090        & 0.070/0.038                   & 0.855/0.657                   \\
BERT\_ab\_facts       & 0.061          & 0.045/0.024                   & 0.774/0.571                   \\
\bottomrule
\end{tabular}
\end{table}

We binarize graded gains $r=\mathrm{wJacc}(A,B)$ at thresholds $\tau\in\{0.20,0.28\}$ to obtain precision–style metrics. 
Raising $\tau$ makes relevance more stringent, so P@k and HitRate@k drop as expected, while the relative model ordering is preserved. 
Across both thresholds and at $k\!\in\!\{10,20,100\}$, the OpenAI embedder decisively outperforms BERT model at both settings. 
At $k{=}10$, for example, P@10 moves from 0.260 to 0.168 and HitRate@10 from 0.716 to 0.550 as $\tau$ rises from 0.20 to 0.28, whereas nDCG@10 (graded) remains at 0.190. 


\begin{table}[ht]
\caption{Paired bootstrap over 2{,}000 common queries (B = 2000 resamples). $\Delta$ is mean difference A$-$B; 95\% CI in brackets. All $p<0.001$.}
\label{tab:bootstrap-k10}
\centering
\setlength{\tabcolsep}{4pt}\footnotesize
\begin{tabular}{llccc}
\toprule
Comparison & Metric & $\Delta$ & 95\% CI & $p$ \\
\midrule
\multicolumn{5}{l}{\emph{$\tau=0.20$}} \\
OpenAI-embedder\_facts $-$ BERT\_ab\_all & nDCG@10 & \textbf{+0.099} & [0.093, 0.105] & $<\!0.001$ \\
OpenAI-embedder\_facts $-$ BERT\_ab\_all & P@10    & \textbf{+0.154} & [0.144, 0.165] & $<\!0.001$ \\
OpenAI-embedder\_facts $-$ BERT\_ab\_all & Hit@10  & \textbf{+0.250} & [0.224, 0.275] & $<\!0.001$ \\
OpenAI-embedder\_facts $-$ BERT\_ab\_facts      & nDCG@10 & \textbf{+0.130} & [0.124, 0.136] & $<\!0.001$ \\
OpenAI-embedder\_facts $-$ BERT\_ab\_facts      & P@10    & \textbf{+0.194} & [0.183, 0.204] & $<\!0.001$ \\
OpenAI-embedder\_facts $-$ BERT\_ab\_facts      & Hit@10  & \textbf{+0.353} & [0.329, 0.378] & $<\!0.001$ \\
\midrule
\multicolumn{5}{l}{\emph{$\tau=0.28$}} \\
OpenAI-embedder\_facts $-$ BERT\_ab\_all & nDCG@10 & \textbf{+0.099} & [0.093, 0.105] & $<\!0.001$ \\
OpenAI-embedder\_facts $-$ BERT\_ab\_all & P@10    & \textbf{+0.108} & [0.099, 0.116] & $<\!0.001$ \\
OpenAI-embedder\_facts $-$ BERT\_ab\_all & Hit@10  & \textbf{+0.236} & [0.211, 0.262] & $<\!0.001$ \\
OpenAI-embedder\_facts $-$ BERT\_ab\_facts      & nDCG@10 & \textbf{+0.130} & [0.124, 0.136] & $<\!0.001$ \\
OpenAI-embedder\_facts $-$ BERT\_ab\_facts      & P@10    & \textbf{+0.130} & [0.121, 0.139] & $<\!0.001$ \\
OpenAI-embedder\_facts $-$ BERT\_ab\_facts      & Hit@10  & \textbf{+0.319} & [0.296, 0.343] & $<\!0.001$ \\
\bottomrule
\end{tabular}
\end{table}

All head-to-head gaps at $k=10$ are statistically significant under paired bootstrap (Table~\ref{tab:bootstrap-k10}), with tight 95\% confidence intervals across both $\tau$ tracks: OpenAI outperforms Bert models on nDCG@10, P@10, and HitRate@10 with $p<0.001$ in every case.

Together with our label-drift analysis, this supports the conclusion that absolute nDCG is conservative under diversified labels, yet relative differences are large and statistically significant.

\subsection{Qualitative Examples}

To illustrate the noisiness of the dataset, we provide the following example of a case law retrieval task. One of the classical issues of criminal law is the right against self-incrimination. The Czech ConC heard many cases on the extent of self-incrimination, with one of the core issues being whether or not the right protects suspects from being subjected to sanctions (e.g., fines) when refusing to cooperate during the collection of DNA samples. ConC consists of individual senates, which on occasion give rise to different answers when ruling on similar legal issues. In this instance, the senates disagreed on the scope of protection and ruled differently (Case no. I. ÚS 671/05 in February 2006, and Case no. III. ÚS 655/06 in May 2007). When the Constitutional court received another petition concerned with the scope of protection, the issue triggered a rare unification procedure (Opinion no. Pl. ÚS-st. 30/10 in November 2010), leading to the establishment of the new scope of protection (Case no. II. ÚS 2369/08 in December 2010). Nowadays, ConC mostly follows the established practice (e.g., Case no. IV. ÚS 666/19 in May 2019), offering little to no distinction. Case law retrieval on the scope of right against self-incrimination should (in both educational and professional settings) include these five documents.\footnote{These five do not exhaust the topic fully, but provide for an example of the noisiness, which is well-known within the Czeh legal community.} We computed $\mathrm{wJacc}$ in Table \ref{keywords} to demonstrate the disparity between keywords assigned in the official court database to these five documents.

\begin{table}[ht]

\centering
\caption{Keyword similarity illustrating noisiness of the keywords assigned to decisions (case numbers abbreviated)}
\label{keywords}
\small
\begin{tabular}{l r r r r r}
\toprule
 & 671/05 & 655/06 & 30/10 & 2369/08 & 666/19 \\
\midrule
671/05 & \cellcolor[HTML]{FF4040}1 & \cellcolor[HTML]{66D070}0.2 & \cellcolor[HTML]{A0E25E}0.3333 & \cellcolor[HTML]{40C47C}0.1111 & \cellcolor[HTML]{66D070}0.2 \\
655/06 & \cellcolor[HTML]{66D070}0.2 & \cellcolor[HTML]{FF4040}1 & \cellcolor[HTML]{A0E25E}0.3333 & \cellcolor[HTML]{7CD669}0.25 & \cellcolor[HTML]{66D070}0.2 \\
30/10 & \cellcolor[HTML]{A0E25E}0.3333 & \cellcolor[HTML]{A0E25E}0.3333 & \cellcolor[HTML]{FF4040}1 & \cellcolor[HTML]{A0E25E}0.3333 & \cellcolor[HTML]{4DC877}0.1429 \\
2369/08 & \cellcolor[HTML]{40C47C}0.1111 & \cellcolor[HTML]{7CD669}0.25 & \cellcolor[HTML]{A0E25E}0.3333 & \cellcolor[HTML]{FF4040}1 & \cellcolor[HTML]{40C47C}0.1111 \\
666/19 & \cellcolor[HTML]{66D070}0.2 & \cellcolor[HTML]{66D070}0.2 & \cellcolor[HTML]{4DC877}0.1429 & \cellcolor[HTML]{40C47C}0.1111 & \cellcolor[HTML]{FF4040}1 \\
\bottomrule
\end{tabular}
\end{table}

\section{Discussion}


Despite being trained on \(\sim\)30k in-domain decisions, our from-scratch BERT lags behind the off-the-shelf OpenAI embedder.
The likely drivers are scale and objective: the general embedder benefits from \emph{orders of magnitude} more pretraining tokens and a broader semantic curriculum, yielding representations that are better aligned for cross-document similarity.
By contrast, our domain model is capacity-limited and trained with MLM only, without contrastive or retrieval-aware signals and without cross-document positives.
Sliding windows (512 with overlap) increase \emph{exposure} to long decisions but do not inject \emph{new} supervision; many windows in a single case are semantically redundant, so averaging them can dilute discriminative cues.
Empirically, the general embedder retrieves neighbors with more and more diagnostic tag overlap even under strict thresholds, indicating that scale and general semantics dominate domain-only MLM at this task.

Absolute nDCG values should be read as \emph{conservative lower bounds} rather than poor utility.
Our gold standard is exact keyword overlap drawn from a large, evolving vocabulary with few tags per decision; as shown in our label-drift diagnostics and overlap model, the expected intersection between two semantically similar decisions is small (\(\mathbb{E}[|A\cap B|]\!\ll\!1\) when \(k^2/N_{\text{eff}}\) is small). We further supported this conclusion with an illustrative qualitative analysis of decisions concerning the same legal issue.
This depresses graded gains even when retrieved decisions are genuinely related.
Head user metrics (P@k, HitRate@k, RBP) and bootstrap gaps nevertheless show large, statistically robust improvements for the OpenAI embedder across \(k\in\{10,20,100\}\) and \(\tau\in\{0.20,0.28\}\).

First, \emph{model choice matters}: a strong general-purpose embedder can outperform a domain-specific BERT trained only with MLM, even on legal texts.
Second, to raise absolute effectiveness under diversified labels, future work should move beyond single-vector document representations toward paragraph-level or multi-vector retrieval, and explore hybrid sparse+dense pipelines (lexical recall plus dense ranking).
We intentionally did no re-ranking here to isolate embedding quality; adding a shallow cross-encoder/listwise re-ranker is a natural next step.

\section{Conclusion and Future Work}

We presented a compact, robust evaluation of semantic retrieval for Czech Constitutional court decisions under noisy institutional labels and found that a strong general-purpose embedder decisively and significantly outperforms a carefully engineered, from-scratch domain BERT trained on approximately 30{,}000 cases. These results suggest that pretraining scale and breadth of semantic coverage matter more than in-domain MLM alone for this task. Looking ahead, we plan to improve absolute effectiveness by moving to paragraph-level (multi-vector) representations and hybrid sparse+dense retrieval, to explore larger-scale continued pretraining of the domain model, and to complement the overlap-based gold with a human assessment by judges and judicial assistants.



\bibliographystyle{vancouver}
\bibliography{bibliography}

\end{document}